\documentclass[letterpaper, 10 pt, conference]{ieeeconf}  

\usepackage{subfigure}
\usepackage{color}
\usepackage{graphicx}   
\usepackage{algorithm}
\usepackage{algorithmic}
\usepackage{amsmath}
\usepackage{amsmath}  
\usepackage{amssymb}  
\usepackage{xcolor} 
\usepackage{tcolorbox}

\definecolor{customblue}{HTML}{dbeef3}
\definecolor{customgreen}{HTML}{80ba8a}
\definecolor{custompurple}{HTML}{c6b3d3}

\IEEEoverridecommandlockouts                              

\title{\LARGE \bf
PAVLM: Advancing Point Cloud based Affordance Understanding 
Via Vision-Language Model
}

\author{Shang-Ching Liu$^{1}$, Van Nhiem Tran$^{2}$, Wenkai Chen$^{1*}$, Wei-Lun Cheng$^{3}$, Yen-Lin Huang$^{4}$, I-Bin Liao$^{2}$,\\ Yung-Hui Li$^{2}$, Jianwei Zhang$^{1}$
\thanks{$^{1}$Technical Aspects of Multimodal Systems (TAMS), Department of Informatics, University of Hamburg}
\thanks{$^{2}$Hon Hai Research Institute (HHRI)}%
\thanks{$^{3}$Department of Electrical Engineering, National Taiwan University}
\thanks{$^{4}$Department of Computer Science and Technology, National Tsinghua University}
\thanks{*Corresponding author to provide e-mail: wchen@informatik.uni-hamburg.de}%
}

\begin{document}

\maketitle
\thispagestyle{empty}
\pagestyle{empty}

\begin{abstract}

Affordance understanding, the task of identifying actionable regions on 3D objects, plays a vital role in allowing robotic systems to engage with and operate within the physical world. Although Visual Language Models (VLMs) have excelled in high-level reasoning and long-horizon planning for robotic manipulation, they still fall short in grasping the nuanced physical properties required for effective human-robot interaction. In this paper, we introduce PAVLM (Point cloud Affordance Vision-Language Model), an innovative framework that utilizes the extensive multimodal knowledge embedded in pre-trained language models to enhance 3D affordance understanding of point cloud. PAVLM is an approach to integrates a geometric-guided propagation module with hidden embeddings from large language models (LLMs) to enrich visual semantics. On the language side, we prompt Llama-3.1 models to generate refined context-aware text, augmenting the instructional input with deeper semantic cues. Experimental results on the 3D-AffordanceNet benchmark demonstrate that PAVLM outperforms baseline methods for both full and partial point clouds, particularly excelling in its generalization to novel open-world affordance tasks of 3D objects. For more information, visit our project site: pavlm-source.github.io.


\end{abstract}

\section{INTRODUCTION}

Affordance understanding is a critical challenge in computer vision, as it is also fundamental to enabling robots to interact effectively with their environment. This understanding is crucial for a wide range of robotic tasks, including human-object interaction~\cite{chen2024ehoa,jian_affordpose_2023}, spatial localization~\cite{Kumar2018VisualMF}, and most importantly, object manipulation~\cite{Bahl2022HumantoRobotII}. Traditional methods often rely on supervised learning from extensive human annotation, which can be time-consuming and limited to specific scenarios~\cite{Li2023LOCATELA}. Moreover, these approaches struggle to generalize to new objects and unseen environments, a critical limitation in robotic manipulation tasks where flexibility and adaptability are essential~\cite{Tremblay2018DeepOP}. For robotic manipulation, affordance understanding involves more than recognizing an object’s 3D geometry, which requires understanding the object's functional properties, specifically how it can be grasped, manipulated, or used in different contexts.  A holistic concept of affordance understanding in real world environments should also take into account the embodiment of the robot, the configuration of the object in its surroundings, and the intended actions of the agent~\cite{ardon2020affordances}. In this work, we mainly focus on the affordance understanding of object point clouds in both full shape and partial shape, which potentially serve as a solution for real-world embodied manipulation.

Despite recent advancements in Visual Language Models (VLMs) from popular open-source models such as LLaVA~\cite{Liu2023VisualIT}, Blip-2~\cite{Li2023BLIP2BL}, to recent open-source SOTA models like Qwen-VL~\cite{Bai2023QwenVLAF}, Phi-3.5-vision~\cite{Abdin2024Phi3TR}, Pixtral~\cite{mistral_pixtral_2024}, and many others. These VLM models, while demonstrating strong visual understanding and reasoning capabilities, remain largely confined to 2D visual tasks involving RGB images. They lack the geometric depth required for effective manipulation in 3D affordance-based tasks. Qian et al.~\cite{qian_affordancellm_2024} were the pioneers in proposing the use of large language models (LLMs) to predict 2D affordance maps by inputting RGB images alongside pseudo-depth information, achieving impressive grounding performance. However, the absence of true depth data limits the robot's ability to interact effectively in 3D space, where a detailed understanding of the physical structure of an object and its possibilities of interaction is critical~\cite{Sawatzky2017WeaklySA}. Consequently, while VLMs demonstrate strong abilities in high-level reasoning and following instructions, their application to robotic manipulation remains limited due to their shallow understanding of the physical world.

\begin{figure}[!t]
\centering
\includegraphics[width=0.48\textwidth]{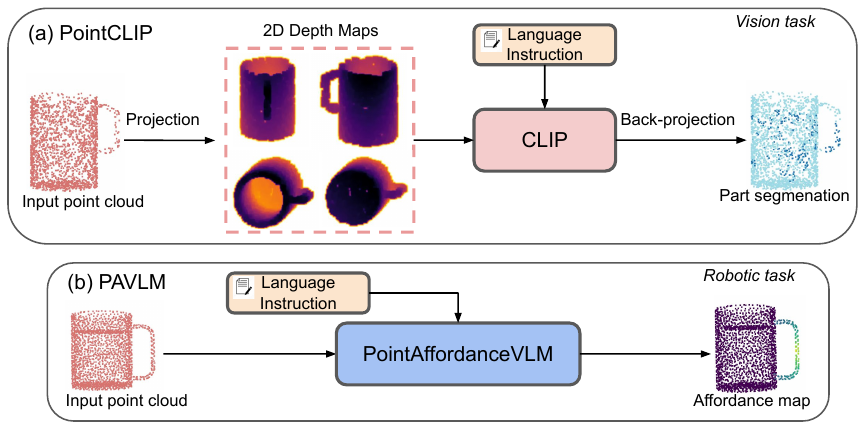}
\caption{Comparison of the overview of (a) PointCLIP series and (b) our PAVLM. PointCLIP series methods rely on a pre-trained VLM, such as CLIP, to process language instructions and multi-view depth maps generated from the point cloud through a series of operations like quantization, densification, smoothing, and squeezing. These complex projection-based steps are limited by the number of views, restricting the model's capacity for comprehensive open-world point cloud understanding. In contrast, our method eliminates the need for these projection steps by employing a geometric-guided point encoder, which directly extracts rich semantic features from the point cloud. Additionally, we fine-tune only a small set of LLM embeddings, maintaining a robust and detailed spatial representation. This approach enhances the understanding of the point cloud from an affordance perspective, providing a more efficient and scalable solution for 3D object interaction.}
\label{Overview}
\vspace{-0.5em}
\end{figure}

To overcome these limitations, some researchers have proposed the use of VLMs to process point cloud data by projecting it onto depth maps, as demonstrated by PointCLIP~\cite{zhang2022pointclip} and PointCLIP V2~\cite{zhu2023pointclip}, shown in Figure~\ref{Overview}. However, these methods involve multiple projection steps, such as quantization, densification, smoothing, and squeezing, which inevitably lead to a loss of geometric and spatial information.  ULIP~\cite{xue2023uliplearningunifiedrepresentation} and ULIP2~\cite{xue2024ulip} have tried to align the point cloud with the image and text that is integrated in LLM as 3D-LLM~\cite{hong20233dllminjecting3dworld} and which will be able to perform the point cloud understanding or point cloud generation task. In this paper, we introduce a framework that integrates knowledge from large language models (LLMs) with 3D point cloud data to enhance point-to-point affordance understanding. We prompt Llama-3.1 models to generate refined, context-aware text, thereby augmenting instructional input with deeper semantic cues. Experimental results on the customized 3D-AffordanceNet benchmark demonstrate that our approach significantly improves performance on open-vocabulary affordance learning tasks and exhibits strong generalization capabilities, particularly in scenarios involving novel objects and unseen environments.

The main contributions of this paper can be summarized as:
\begin{itemize}
\item  We propose a unified architecture that integrates 3D point cloud data with an open-source language model for point cloud affordance understanding, especially considering the real-robot partial views.
\item We introduce a comprehensive augmented data pipeline for both 3D point cloud objects and corresponding question-answer pairs for each training sample.
\item Our approach outperforms state-of-the-art methods on the 3D-AffordanceNet benchmark, demonstrating a stronger generalization to novel object categories.
\end{itemize}
\section{RELATED WORK}

\textbf{Supervised Learning for Affordance Understanding}
As affordance understanding in the robotic field, it can be simply regarded as object detection and semantic segmentation tasks as a labeling process~\cite{hassanin_visual_2022}. However, when a robotic agent interacts with objects, it must recognize them not only by their physical characteristics but also by the actions they enable. Thus, a robust understanding of affordances should improve a robot's ability to perform appropriate actions tailored to diverse tasks and human requirements. With the rise of deep learning, several methods have emerged that leverage neural networks for learning visual affordance. Notably,~\cite{7759429} introduced an encoder-decoder architecture designed to predict affordance labels from RGB images. Building on this,~\cite{8620534} further advanced affordance detection and segmentation in real, unlabeled data by learning from synthetic RGB datasets. To incorporate richer geometric information, benchmarks such as 3D AffordanceNet~\cite{do_affordancenet_2018} and PartAfford~\cite{xu_partafford_2022} were developed, enabling more sophisticated 3D affordance reasoning. These datasets provide dense per-point affordance annotations, illustrating the spatial distribution of interaction regions in 3D space. Moreover, recent advances have inspired context-aware affordance applications such as task-oriented grasp estimation~\cite{9981900} and human hand pose detection~\cite{jian_affordpose_2023}. Despite these advancements, most prior work on visual affordance relies on supervised or self-supervised learning, limiting the model's ability to generalize to novel objects or affordance labels. Our proposed PAVLM overcomes these limitations by harnessing the zero-shot capabilities of pretrained vision language models. By utilizing 3D point cloud data in conjunction with pre-trained language models, our approach significantly extends generalization to unseen object categories. 

\textbf{Pretraining from 3D Point Clouds}
To effectively obtain feature representations from point clouds,  PointNet++~\cite{qi_pointnet_2017} introduced a hierarchical approach to progressive group neighbouring points and learned local features at multiple scales. Drawing inspiration from the concept of mask-based pretraining, Point-MAE~\cite{pang_masked_2022}  and Point-FEMAE~\cite{zha_towards_2023} introduced a masked
auto-encoding scheme to prevent leakage of location information and uneven information density of the point cloud feature. They significantly boost performance in tasks like point cloud reconstruction tasks. 
Inspired by the success of BERT~\cite{devlin2018bert} in natural language processing, Point-BERT~\cite{Yu2021PointBERTP3} adopted the transformer-based Masked Point Modeling (MPM) scheme for point clouds. This method randomly masks out patches of point cloud data and feeds the masked input into a transformer backbone for pre-training. Although these approaches improved point cloud understanding, experiments demonstrated its limited capacity to generalize across new tasks and domains.

The success of transformers in large language models (LLMs) has opened the door to incorporating point cloud features into LLMs. Point-Bind LLM~\cite{guo_point-bind_2023}, based on ImageBind~\cite{girdhar_imagebind_2023} work, connects six different modalities, including text, images, audio, video, IMU, and depth data, achieving a more comprehensive multimodal understanding. To further enhance point cloud embeddings, ImageBind-LLM~\cite{han_imagebind-llm_2023} provided instruction-based fine-tuning after alignment, enabling the LLM to better interpret inputs and generate more accurate outputs for specific tasks. This process leverages the LLaMA-Adapter~\cite{zhang_llama-adapter_2023}, introducing an adapter module to fine-tune LLMs for specialized tasks like point cloud processing. Our approach adopts a similar pretraining strategy to leverage open-world knowledge from the pretrained multimodal LLMs, but specifically adapts it to a robot-oriented setting focused on point cloud-based affordance understanding.

\begin{figure*}[!t]
  \centering
  \includegraphics[width=\textwidth]{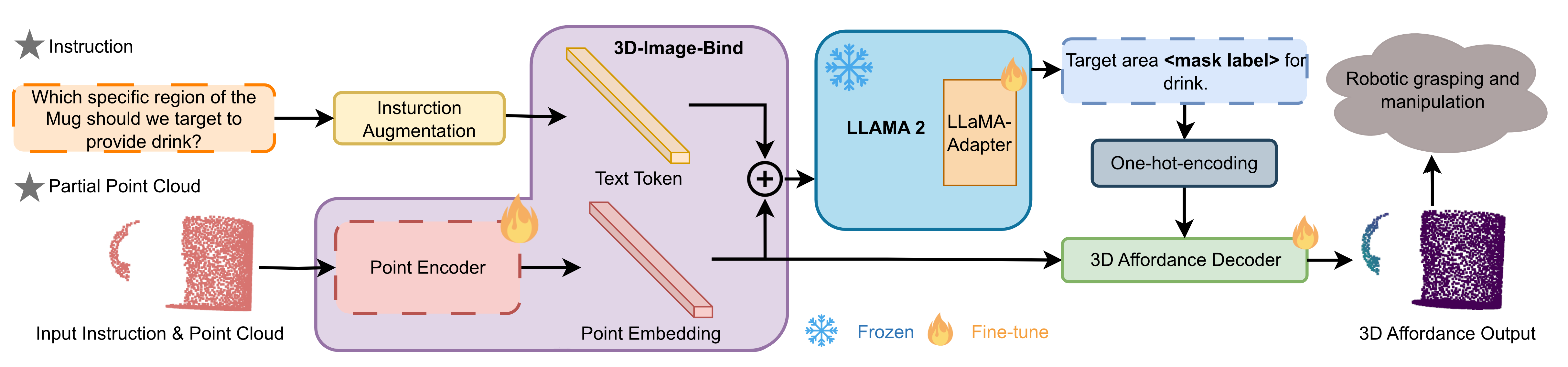}
  \caption{The pipeline of our PAVLM. To encode point cloud data directly, we devise a geometric-guided propagation module to obtain point features. On the language side, we enhance textual instructions by prompting Llama-3.1 to generate richer and more detailed guidance. Both visual and textual embeddings are processed through the 3D Image-Bind approach to achieve feature alignment, and the concatenated features are fed into a multi-modal LLM (Llama-2) for mask label generation. Finally, the per-point feature embeddings, derived from the geometric-guided propagation module, are multiplied by the embedding corresponding to the \texttt{<mask label>} token, which is then input into a 3D affordance decoder to generate the corresponding affordance map.}
  \label{fig:framework}
\end{figure*}
\textbf{Visual Language Model for Manipulation }
The emergence of LLM-based intelligent agents with visual understanding, such as LLAVA~\cite{liu_visual_2023} and ImageBind-LLM~\cite{han_imagebind-llm_2023}, has significantly advanced the field of Embodied AI (E-AI)~\cite{paolo_call_2024}. Visual Language Models (VLMs) have been particularly effective in enabling high-level, long-horizon planning for robotic tasks, leveraging the reasoning capabilities of LLMs. For example, Voxposer~\cite{huang_voxposer_2023} combines VLMs with the code-as-policy approach~\cite{liang_code_2023} to generate Python code for robotic manipulation tasks. However, this approach is somewhat simplistic, often focusing on locating an object's midpoint without considering critical affordances, such as grasping a long object like a knife by the handle instead of the middle. 

To address this limitation, Copa~\cite{huang_copa_2024} introduced the Segment Anything Model (SAM), which allows for more precise selection of grasping points, enhancing the generalization abilities of Voxposer. However, since SAM operates only in 2D, it limits planning for robotic manipulation in 3D space. While 2D affordance maps are commonly used, AffordanceLLM~\cite{qian_affordancellm_2024} was the first to integrate a 2D affordance map with an LLM to improve the agent’s reasoning abilities. Despite this, the gap between 2D affordance reasoning and real-world 3D manipulation remains a challenge. To address this, approaches such as LASO~\cite{Li_2024_CVPR} have leveraged language output as cues for 3D point cloud feature reconstruction. However, these methods do not fully exploit the multimodal capabilities of pre-trained large language models that are already aligned with point cloud data. To overcome this limitation, we propose PAVLM, a framework designed to generate 3D object affordances, potentially enabling more effective robotic manipulation in real-world environments.

\section{The proposed PAVLM Framework}

In this section, we present our approach, which takes as input a point cloud $P$ and a textual instruction
$T$ to produce a 3D affordance map guided by a \texttt{<mask label>}. The mask label is used to integrate with point cloud embeddings to accurately generate the affordance map for the point cloud. 

More detailed, the process begins with the geometric-guided point encoder, which extracts geometric features from the input point cloud. These features are then infused into the point cloud embedding to enrich the representation. Next, we utilize the Llama-3.1 LLM model through prompting to enhance the reasoning capabilities of the system via data augmentation, generating more diverse context-aware question-answer pairs. Afterward, we adopt a pre-trained 3D Image-Bind model to align the point cloud embedding with the textual tokens, ensuring multimodal consistency between the 3D data and the instruction. Subsequently, the fine-tuned Llama-2 model generates affordance mask labels, which we integrate with the point cloud embeddings. We selected Llama-2 due to the limitations of LLaMA-Adapter v1 \cite{zhang2023llama}, while still benefiting from its ability to incorporate point-cloud knowledge, as seen in ImageBind-LLM\cite{han_imagebind-llm_2023}. Furthermore, pre-trained Llama-2 knowledge supports our affordance task sufficiently without requiring extensive language-specific fine-tuning. Thus, we focus primarily on fine-tuning it for affordance map label classification. Finally, these embeddings are fed into a 3D affordance decoder, which produces the final affordance map with high precision. A detailed pipeline of our framework is shown in Figure~\ref{fig:framework}. 

\begin{figure}[h!]
\centering
\includegraphics[width=\linewidth]{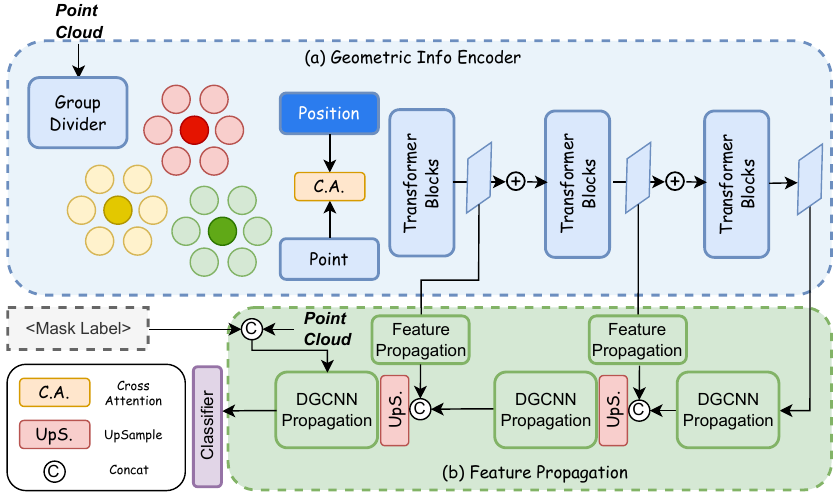}
\caption{Architecture of the proposed Geometric-guided Point Encoder in our PAVLM pipeline.
}
\label{framework_detail}
\end{figure}
\subsection{Geometric-guided Point Encoder}

As illustrated in Figure~\ref{framework_detail}, we propose a geometric-guided point encoder to generate the final point cloud embeddings. This encoder is composed of two key submodules: (a) the geometric information extraction module and (b) the feature propagation module. In the first module, we begin by dividing the input point cloud \( P \in \mathbb{R}^{2048 \times 3} \) into \( k \) point patches. Inspired by Point-BERT~\cite{Yu2021PointBERTP3}, we define a constant K to form K groups, each containing 256 points. We then apply convolutional layers to these patches to extract point-wise features \( \{ f_i \}_{i=1}^{k} \). The features are added to the positional embeddings \( \{ p_i \}_{i=1}^{k} \) and fed into a series of transformer blocks via a cross-attention mechanism. This process yields refined geometric features \( \{ g_i \}_{i=1}^{k} \). We adopt a structure comprising 3-layer transformer blocks, and the extraction of geometric features can be formulated as follows:
\begin{equation}
g_f = \mathcal{G}(f_i, p_i), \quad 
\hat{F}_j = F_j + g_j \cdot \text{softmax} \left( \frac{F_j g_f^T}{\sqrt{D}} \right) g_f,
\end{equation}
where \( F_j \) represents the output feature from the \( j \)-th transformer layer. To enhance the quality of point embeddings, we incorporate a feature propagation module, as recommended by~\cite{he2024segpoint}. Specifically, we process the output features to generate high-quality per-point embeddings. We employ an upsampling technique to propagate the features from the first and second transformer block outputs, whereas Dynamic Graph CNN (DGCNN~\cite{phan2018dgcnn}) propagation techniques are also used to ensure the features are effectively distributed across the point cloud.



\subsection{Textual Instruction Augmentation}
\label{sec:text_aug}
To enhance affordance understanding through the alignment of point cloud data and text, we incorporate 3D-specific descriptions that reflect implicit human-object interaction characteristics as the textual input. Rather than using general action terms like cut, handover, and tool use, commonly seen in previous affordance-based robotic works~\cite{9981900,nguyen2024language}, we adopt a more detailed question-answering approach inspired by AffordanceLLM~\cite{qian_affordancellm_2024}. In this approach, a seed question is generated, such as ``What part of the \texttt{<object\ name>} should we interact with to \texttt{<action\ name>} it?'' followed by an answer like ``You can \texttt{<action\ name>} the area \texttt{<mask\ token>}''.

\begin{figure}[htbp]
\vspace{-0mm} 
\centering
\includegraphics[width=0.99\linewidth]{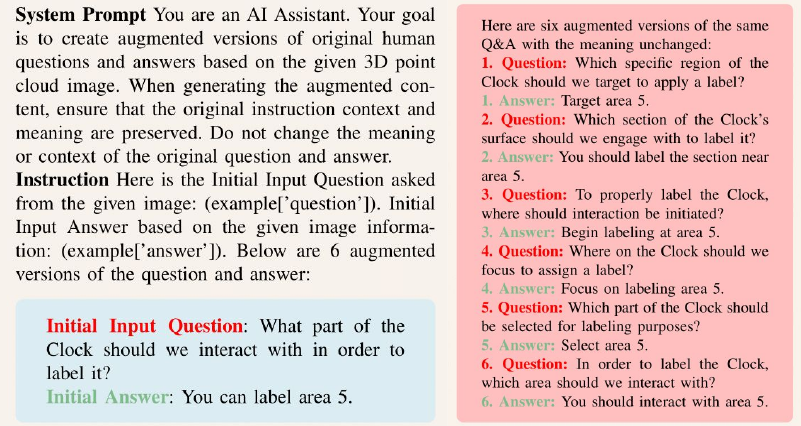}
\caption{The augmentation prompt template of our textual instructions to enhance the diversity of question and answer pairs.
}
\label{prompt}
\end{figure}

Given the powerful descriptive capabilities of large language models (LLMs), we extend this approach by using Llama-3.1 70B~\cite{llama3modelcard} to generate richer, 3D-specific augmentation question-answering pairs. Llama-3.1 processes the language command and outputs responses based on its pre-trained knowledge, resulting in more detailed and context-aware interactions. An example of the prompt we used for this process is illustrated in Figure~\ref{prompt}. For a 3D dataset with $K$ categories, we replace the \texttt{<mask label>} in each command with the corresponding category value and input these commands into Llama-3.1, generating 3D-specific descriptions enriched with category-specific semantics.

\subsection{Feature Alignment of Point Cloud and Text}
\label{sec:alignnment}
Inspired by Image-Bind~\cite{girdhar_imagebind_2023}, it integrates five modalities, including video, depth, IMU, audio, text, and thermal data to achieve cross-model feature learning. Point-Bind~\cite{guo_point-bind_2023} extends this idea by constructing a joint embedding space between 3D data and multiple modalities. After training on large-scale point-paired data, Point-Bind effectively aligns point cloud data with images, text, and audio into a unified representation space. By introducing a contrastive loss function, Point-Bind achieves 3D-centric multimodal learning. 
Building on this idea,  we align the extracted feature from the point cloud $P$ and text $T$ with the contrastive batch average loss \( L_c \) by calculating their Euclidean distance and maintain the minimum margin as 1, shown in Equation ~\ref{eq:contrastive_loss}.

\begin{equation}
\label{eq:contrastive_loss}
    \begin{split}
    L_{ca} = \text{Average}(\text{dist}(P, T)) \\
    =\text{Average} \| P - T + \textit{eps} \cdot e \|_p ,
    \end{split}
\end{equation}

where  $\textit{eps} \cdot e$ the term is used to avoid numerical instability or division by zero in distance calculations.

\subsection{Masked-based Affordance Decoding}
Continually, the aligned feature embeddings generated by 3D Image-Bind are fed into a pretrained multimodal Llama-2 model. In this process, we freeze most of the Llama-2 weight parameters and apply the fine-tuning strategy of the LLaMA-Adapter~\cite{zhang_llama-adapter_2023} to predict a special \texttt{<mask label>}. Based on the 3D AffordanceNet benchmark~\cite{deng_3d_2021}, we define 18 categories of affordance mask labels. To encode the output labels, we employ a one-hot encoding scheme, transforming the hidden state of the \texttt{<mask label>} into a query embedding \( q \).

Seen in Figure~\ref{framework_detail}, this query embedding \( q \) first concatenates with the original point cloud $P$, and then integrated with point cloud embedding  \( P_{em} \) into the 3D affordance decoder, which consists of two convolutional layers and a batch normalization layer. Its primary function is to reduce the dimensionality of the multimodal feature embeddings and produce a point-wise affordance map \( \mathbf{M} \in \mathbb{R}^{2048 \times 1} \) in the point cloud, where each point \( M_i \in [0, 1] \) represents the affordance score for a point \( i \) in the point cloud. The above process can be mathematically expressed as:
\begin{equation}
\label{eq:mask_encoding}
    \mathbf{M} = \text{Decoder}(P_{em}, (P \oplus q)),
\end{equation}
\begin{equation}
\mathbf{M} = \{ M_i \mid M_i \in [0, 1], \, i = 1, 2, \dots, 2048 \}.
\end{equation}

\subsection{Training Objectives}

For the comprehensive training of both the geometric-guided encoder and the 3D affordance decoder, we adopt a multi-objective loss function that integrates several components to ensure effective learning. The core affordance loss \(L_{aff}\) is composed of a cross-entropy loss \(L_{ce}\) and a Dice loss \(L_{dice}\), following the practice established in prior work~\cite{deng_3d_2021}. This combination helps to optimize both the pixel-wise classification and spatial consistency of the affordance predictions. In addition to the affordance loss, we introduce an alignment loss to ensure that the point cloud embeddings are well aligned with the textual modality, as defined in Eq.~\ref{eq:contrastive_loss}. This alignment is crucial for integrating 3D geometric information into the joint representation space. Furthermore, when fine-tuning the Llama-2 model with LLaMA-Adapter, we introduce an additional query loss \(L_q\), which is a cross-entropy loss aimed at improving the learning of affordance labels based on the query embeddings generated from the textual prompts.  The overall affordance learning loss during the training process is defined as:
\begin{equation}
L_{aff} = L_{ce} +  \lambda \cdot L_{dice},
\end{equation}
where the weighting factor \(\lambda\) is set to 1. Both the alignment loss \(L_{ca}\) and the query loss \(L_q\) are used independently to fine-tune the pre-trained multimodal model.

\section{Experiments}
Given a 3D point cloud of an object, our objective is to estimate affordances for both complete and partial shapes. In our experiments, we systematically evaluate the performance of our approach by addressing two key questions: (1) How do specific design choices, such as varying prompts and visual encoders, affect overall performance? (2) How well does our method generalize in comparison to state-of-the-art techniques?

\subsection{Experiment Setup}
\textbf{Metrics.} We assess affordance estimation using four metrics: mean Average Precision (mAP), Area Under the ROC Curve (AUC), average Intersection Over Union (aIoU), and mean squared error (MSE). Firstly, we generate a Precision-Recall curve and compute the Average Precision for each affordance. The area under the ROC curve is calculated as the AUC metric. We also compute the mean squared error (MSE) for each affordance class and sum the results across all categories. After binarizing the predictions in a step of 0.01, the aIoU is calculated by progressively increasing the threshold from 0 to 0.99. Aside from MSE, all metrics are further processes in an average mathematical operation. All numerical values reported in the tables have been scaled by a factor of 100 (i.e., multiplied by 100) and values are rounded at the last reported digit.

\textbf{Baselines.} We compare our approach with state-of-the-art baseline methods. As shown in Figure~\ref{Overview}, we mainly compare our work with PointCLIP~\cite{zhang2022pointclip}  PointCLIP V2~\cite{zhu2023pointclip}, ULIP~\cite{xue2023uliplearningunifiedrepresentation}, and ULIP2~\cite{xue2024ulip} which all achieve great generalization performance on the language-interactive 3d point cloud understanding tasks. 
\begin{itemize}
\item \textbf{PointCLIP and PointCLIP V2}~\cite{zhang2022pointclip, zhu2023pointclip}: PointCLIP bridges the gap between 2D image-text pre-trained models (CLIP) and the irregular structure of 3D point clouds by introducing a multi-view projection-based method. This approach transforms point clouds into multi-view depth maps, which are then processed by CLIP’s visual encoder, achieving promising results on standard 3D benchmarks such as ModelNet and ScanObjectNN. Building upon this, PointCLIP V2 enhances the framework by integrating CLIP and GPT models for robust 3D open-world learning. It introduces more realistic visual projections and rich 3D-specific text prompts, making it a versatile tool for open-world point cloud understanding.

\item \textbf{ULIP and ULIP-2}~\cite{xue2023uliplearningunifiedrepresentation, xue2024ulip}: ULIP is a multimodal pre-training framework that aligns 3D point clouds, images, and textual descriptions into a unified representation space. By leveraging pre-trained vision-language models, ULIP enhances 3D representation learning, enabling state-of-the-art performance in tasks like zero-shot 3D classification and image-to-3D retrieval. Extending this approach, ULIP-2 employs large multimodal models to automatically generate comprehensive language descriptions for 3D objects, facilitating scalable multimodal pre-training without manual annotations. This advancement further improves 3D understanding and cross-modal applications.

\end{itemize}

\begin{table}[b]
\caption{Ablation study of different text prompts.}
\label{tab:text-prompts}
\resizebox{\linewidth}{!}{%
\begin{tabular}{cccccc}
\hline
Point cloud   & Text Prompt    & mAP$\uparrow$ & AUC$\uparrow$ & aIOU$\uparrow$ & MSE$\downarrow$ \\ \hline
Full-shape (Aug.)& Ours             &  45.7   & 86.9 & 16.8 &  0.45    \\
Full-shape    & Ours           &  \textbf{46.0}   & 87.1 & \textbf{17.1} & \textbf{0.44}     \\
Full-shape    & Object, Action &   44.5  & 85.6 & 16.0 &  0.67     \\
Full-shape    & Action         &   45.4  & \textbf{87.2}  & 16.7&  0.45   \\
Full-shape    & Hi             &   11.4  & 51.9 & 0.4&  0.90\\
\hline

\end{tabular}}
\end{table}
\textbf{Dataset.} We primarily evaluate our methods using the 3D AffordanceNet benchmark~\cite{deng_3d_2021}, which is specifically designed for visual affordance understanding in 3D. The dataset contains 22,949 3D object models across 23 semantic categories and covers 18 affordance categories. During the training process, we employ two dataset-splitting strategies: \textit{seen} and \textit{unseen}. For the \textit{seen} strategy, we shuffle the original dataset and split it into training, validation, and test sets using an 8:1:1 ratio where each of these splits contains all categories. In contrast, the \textit{unseen} strategy follows the same splitting ratio but introduces specific object categories, namely ``mug,'' ``knife,'' and ``scissors'', which are entirely reserved for the test set. These categories are excluded from the training and validation sets, making this split particularly valuable for evaluating the model's ability to generalize to novel objects, which is significant for real-world robotic manipulation tasks. We have further processed the data into rows containing the fields ``instruct'', ``input'', ``Answer'', and ``Affordance\_map'' to facilitate downstream tasks. The original template for the ``instruct'' field, inspired by AffordanceLLM~\cite{qian2024affordancellm}, is formulated as: ``What part of the \{semantic\_class\} should we interact with in order to \{affordance\} it?'' Correspondingly, the ``Answer'' field is structured as: ``You can \{affordance\} the area \{mask\_token\}.'' The ``instruct'' and ``answer'' field is further augmented using the method described in Section \ref{sec:text_aug}.

\subsection{Ablation Study}
We performed several ablation studies to evaluate how various modules of our model contribute to its overall performance on the \textit{seen} split dataset. Specifically, we examined the effects of varying text prompts and point cloud encoders. Following the evaluation prompts used in previous work~\cite{jian_affordpose_2023}, we tested four different types of text prompts to analyze their impact on model performance: (1) A simple greeting noise, such as ``Hi'', (2) The action associated with the object, (3) A combination of the object name and action label, and (4) A full question prompt and its augmentation, such as ``Which specific region of the mug should we target to provide drinking?''. 

The results, summarized in Table~\ref{tab:text-prompts}, indicate that comprehensive question prompts consistently outperform simpler ones across all four metrics. Notice that when the context closely matches the original dataset prompt, better results are produced. Compared to object names, action names combined with point cloud embeddings provide stronger hints toward the result, indicating that object names are less correlated with the outcome than their associated actions. Furthermore, when textual enhancement is applied, the performance metrics are slightly lower than those of the original prompts. This is expected, as the augmented dataset enhances the LLM's generalization capability. Although the exact sentence text yields better results, the model becomes more adept at handling complex language instructions.



In addition, we compared our geometric-guided point encoder against two widely used point cloud-based network architectures: PointNet++~\cite{qi_pointnet_2017} and DGCNN~\cite{phan2018dgcnn}. As demonstrated in Table~\ref{tab:vision-encoder}, our proposed geometric-guided point encoder average outperforms them when used as the vision backbone. For instance, in key metrics like mAP and aIoU, our model achieves over 10\% higher performance than the best PointNet++ model. The main reason is that feature propagation provides sufficient cues to the upsampling mechanism, especially in partial shapes, so that before classification, there is already a strong understanding of the object. The framework allows the mask label to trigger the activation part more effectively, resulting in better performance. This enhanced performance indicates that our encoder is more effective in extracting meaningful semantic feature embeddings for our affordance understanding task.

\begin{table}[!t]
\caption{Ablation study of different vision encoders.}
\label{tab:vision-encoder}
\begin{tabular}{cccccc}
\hline
Point cloud   & Vision encoder & mAP$\uparrow$ & AUC$\uparrow$ & aIOU$\uparrow$ & MSE$\downarrow$ \\ \hline
Full-shape    & PointNet++        &    47.1 &   86.2  &   \textbf{18.1}  &  \textbf{0.36}   \\
Full-shape    & DGCNN          &  31.0   &   72.9  &  1.0   &  0.68   \\
Full-shape    & Ours           &  \textbf{48.5}&  \textbf{86.8}   &  17.7   & \textbf{0.36}    \\
\hline
Partial-shape & PointNet++        &   25.3  &  69.8   &  3.4   &  0.73   \\
Partial-shape & DGCNN          &     26.8& 69.7    &  0.7   &   0.67  \\
Partial-shape & Ours           &  \textbf{40.2}  &   \textbf{82.2}  &  \textbf{11.7}   &  \textbf{0.39}   \\ \hline
\end{tabular}
\end{table}

\subsection{Evaluation Result}

\textbf{Comparison with Baselines}.
To further evaluate 3D affordance understanding, we compared the performance of our model with state-of-the-art point cloud understanding approaches, specifically PointCLIP~\cite{zhang2022pointclip}, PointCLIP V2~\cite{zhu2023pointclip}, ULIP~\cite{xue2023uliplearningunifiedrepresentation}, and ULIP2~\cite{xue2024ulip}. Since all baseline methods leverage the frozen pre-trained vision-language model such as PoinCLIP from CILP and ULIP have reported using Pointbert as their core point encoder would perform the best, we introduced a variant of our model by freezing the pre-trained Image-Bind and Llama-2 modules, then further fine-tune by our dataset. It is noted that previous baseline methods typically achieve point cloud understanding using full-shape point clouds from a computer vision perspective. In contrast, we emphasize real-robot visual understanding by introducing extra partial-shape point clouds. The results presented in Table~\ref{tab:comparison-methods} are obtained using the original methods and their best-released checkpoints, fine-tuned for three epochs, and essential data transformations to fit the input size and model requirements. Note that the PointCLIP and ULIP  series were originally trained on other tasks in their original repository, which we assume gave them general capabilities for understanding the point cloud in downstream tasks. In contrast, our model, including the frozen variant, consistently delivers superior performance across all four metrics, demonstrating its effectiveness on both full-shape and partial-shape point cloud data.

\begin{table}[!t]
\caption{Comparison results with baselines on the \textit{seen} split of dataset}
\label{tab:comparison-methods}
\begin{tabular}{cccccc}
\hline
Point cloud   & Method       & mAP$\uparrow$ & AUC$\uparrow$ & aIOU$\uparrow$ & MSE$\downarrow$ \\ \hline
Full-shape    & PointCLIP    &  7.6   &   49.9  &  0.9   &  0.80   \\
Full-shape    & PointCLIP V2 &   7.6  &  50.0   &   0.8  & 0.70    \\
Full-shape    & ULIP  &  7.6   &  50.0   & 0.4    &  0.60\\
Full-shape    & ULIP2  &  7.6   &  50.1   & 1.2    &  0.63   \\
Full-shape    & Ours (frozen) & 46.3   &  \textbf{86.9}   &  17.1  &  0.44    \\
Full-shape    & Ours         &  \textbf{48.5}   &   86.8  &  \textbf{17.7}   & \textbf{0.36}    \\ \hline
Partial-shape & PointCLIP    &  9.1  &50.1 &  1.1   &     0.78     \\
Partial-shape & PointCLIP V2 &  9.2   &  49.9   &  1.4   &  0.66   \\
Partial-shape & ULIP &  9.2   &  50.0   &  1.0   & 0.58    \\
Partial-shape & ULIP2 &   9.2  &  50.1   &  1.1   &  0.55   \\
Partial-shape & Ours (frozen) &  37.2   &  64.2   & 11.4    &  \textbf{0.36}   \\
Partial-shape & Ours          &  \textbf{40.2}   &   \textbf{82.2}  &  \textbf{11.7}   &  0.39    \\ \hline
\end{tabular}
\end{table}

\textbf{Generalization to Open-world Categories}. To evaluate the generalization capability of our model to open-world object categories, we conducted experiments on the \textit{unseen} split of the dataset, using the same experimental setup. As presented in Table~\ref{tab:generalization-unseen}, the performance of all models, across both full shape and partial shape evaluations, shows a noticeable decline compared to the results on the \textit{seen} split, as reported in Table~\ref{tab:comparison-methods}. Despite this overall drop, our model significantly outperforms all baseline methods, particularly in terms of the mAP and MSE metrics. This highlights its strong ability to generalize to novel object categories, even in the challenging task of affordance understanding.

\subsection{Visualization Result}

\begin{figure}[!t]
  \centering
  \includegraphics[width=\linewidth]{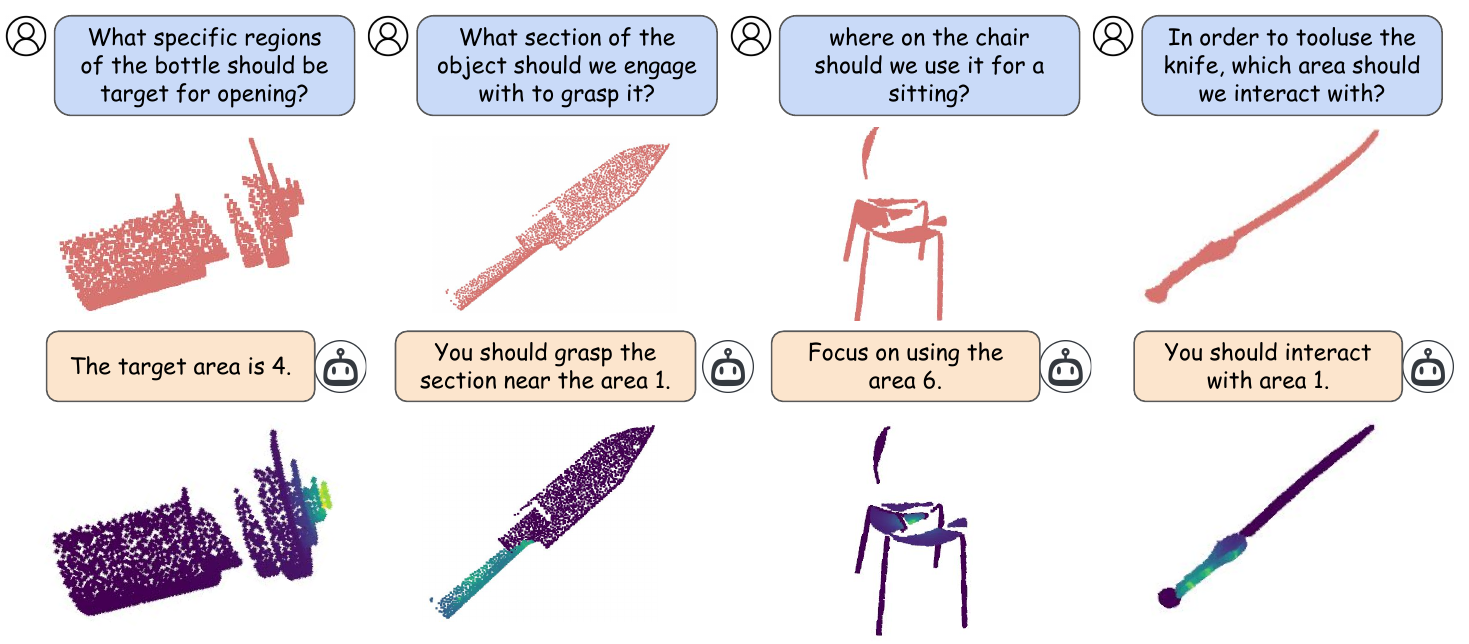}
  \caption{Visualization of affordance prediction results on multiple unseen objects and categories, showing that our model predicts the accurate affordance as different instructions guides from the original partial point cloud.}
  \label{fig: visualization}
\end{figure}

As illustrated in Fig~\ref{fig: visualization}, we present several qualitative results from our model on unseen objects and categories. When given an implicit instruction, such as ``Where on the chair should we use it for sitting?'', our model accurately identifies the corresponding affordance region on the chair. In some categories from the \textit{unseen} split, like knives, more complex instructions that require extensive reasoning ability, such as ``To use the knife as a tool, which area should we interact with?'', are handled effectively by our model. It accurately predicts the handle centre of the knife. These results demonstrate that our method leverages the reasoning capabilities of LLMs to generate high-quality affordance masks, even for challenged partial point cloud understanding.



\section{Quantitative Limitations}

While our proposed framework, PAVLM, demonstrates significant advancements in 3D affordance understanding by integrating geometric-guided propagation with large language models (LLMs), certain quantitative limitations must be acknowledged. Due to time constraints and overlapping research in the visual domain, we were unable to perform direct, method-wise comparisons with similar work, such as LASO~\cite{Li_2024_CVPR} and 3D-LLM~\cite{hong20233dllminjecting3dworld}. However, a theoretical comparison can be explained. 3D-LLM employs an approach similar to that of PointCLIP by extracting multi-view images from the 3D scene, deriving dense 2D features, and concatenating them to reconstruct 3D representations—a process that inevitably leads to some loss of the original 3D information. In contrast, LASO does not leverage 3D embeddings to fully exploit the multimodal capabilities of pre-trained LLMs, which may limit its performance in open-world recognition tasks involving unseen categories compared to PAVLM.

\section{CONCLUSIONS}

We have introduced PAVLM, a novel framework for 3D point cloud-based affordance understanding that successfully handles unseen, open-world objects. Our approach leverages the extensive knowledge embedded in Multimodal Large Language Models (MLLMs) to enhance performance, which significantly surpasses state-of-the-art affordance understanding models, excelling in both full-shape and partial-shape point clouds, even under complex language instructions. This enhanced capability is vital for the development of embodied robots that can adeptly grasp and manipulate objects in dynamic and human-robot collaborative environments. In future work, we plan to explore how to adapt the proposed PAVLM framework to further improve its capabilities and extend its applications to integrate with 6-DOF grasp generation algorithms, enabling its application in real-world robotic tasks.

\begin{table}[!t]
\caption{Comparison results with baselines on the \textit{unseen} split of dataset}
\label{tab:generalization-unseen}
\begin{tabular}{cccccc}
\hline
Point cloud   & Method       & mAP$\uparrow$ & AUC$\uparrow$ & aIOU$\uparrow$ & MSE$\downarrow$ \\ \hline
Full-shape    & PointCLIP    &  3.7   &  49.9   &   0.5  &  1.58   \\
Full-shape    & PointCLIP V2 &  3.7   &  49.3   &  0.4   &  0.91   \\
Full-shape    & ULIP &   3.8  &   50.2  &  0.17   &  0.53\\
Full-shape    & ULIP2 &  3.8   & 49.9    & 0.22    &  0.70\\
Full-shape    & Ours (frozen) &  11.8   &  \textbf{60.3}   &  1.5  &  0.78    \\
Full-shape    & Ours         &   \textbf{12.1}  &  53.9   &  \textbf{2.6}   &  \textbf{0.51}   \\\hline
Partial-shape & PointCLIP    &  4.7  & 50.2 &  0.6   &      1.14    \\
Partial-shape & PointCLIP V2 & 4.7    & 50.3     & 0.38    &  0.93   \\
Partial-shape & ULIP &  4.8   &   50.6   &   0.23  &  0.66   \\
Partial-shape & ULIP2 &   4.8  &  50.0    &   0.37  &  0.61   \\
Partial-shape & Ours (frozen) & 13.8   &  60.3  &  2.0  &   \textbf{0.60}  \\
Partial-shape & Ours         &   \textbf{22.3}  & \textbf{63.2}    &   \textbf{2.1}  &   0.82  \\ \hline
\end{tabular}
\end{table}

\section{Acknowledgements}
This research was funded by the German Research Foundation (DFG)
and the National Science Foundation of China (NSFC)
in the project Crossmodal Learning, DFG TRR-169/NSFC 62061136001, and Hon Hai Research Institute (HHRI).

\bibliographystyle{IEEEtran}
\bibliography{bib/reference}

\end{document}